\documentclass[10pt,twocolumn,letterpaper]{article}

\usepackage{iccv}
\usepackage{times}
\usepackage{epsfig}
\usepackage{graphicx}
\usepackage{amsmath}
\usepackage{amssymb}

\usepackage{subfig}
\usepackage{multirow}
\usepackage{arydshln}
\usepackage{placeins}
\usepackage{here}
\usepackage[breaklinks=true,bookmarks=false]{hyperref}

\iccvfinalcopy % *** Uncomment this line for the final submission

\def\iccvPaperID{0003} % *** Enter the ICCV Paper ID here 

\ificcvfinal\fi

\begin{document}

%%%%%%%%% TITLE
\title{Probabilistic Vehicle Reconstruction Using a Multi-Task CNN}

\author{Max Coenen \qquad Franz Rottensteiner  \\
Leibniz University Hannover, Germany\\
%Institution1 address\\
{\tt\small \{coenen,rottensteiner\}@ipi.uni-hannover.de}
}

\maketitle
\ificcvfinal\thispagestyle{empty}\fi

%====================================================================================================
%%%%%%%%% ABSTRACT
\begin{abstract}
The retrieval of the 3D pose and shape of objects from images is an ill-posed problem. A common way to object reconstruction is to match entities such as keypoints, edges, or contours of a deformable 3D model, used as shape prior, to their corresponding entities inferred from the image. However, such approaches are highly sensitive to model initialisation, imprecise keypoint localisations and/or illumination conditions. In this paper, we present a probabilistic approach for shape-aware 3D vehicle reconstruction from stereo images that leverages the outputs of a novel multi-task CNN. Specifically, we train a CNN that outputs probability distributions for the vehicle's orientation and for both, vehicle keypoints and wireframe edges. Together with 3D stereo information we integrate the predicted distributions into a common probabilistic framework. We believe that the CNN-based detection of wireframe edges reduces the sensitivity to illumination conditions and object contrast and that using the raw probability maps instead of inferring keypoint positions reduces the sensitivity to keypoint localisation errors. We show that our method achieves state-of-the-art results, evaluating our method on the challenging KITTI benchmark and on our own new 'Stereo-Vehicle' dataset.
\end{abstract}

%====================================================================================================
\section{Introduction}
\begin{figure}[t]
\centering
\includegraphics[width=1.0\columnwidth]{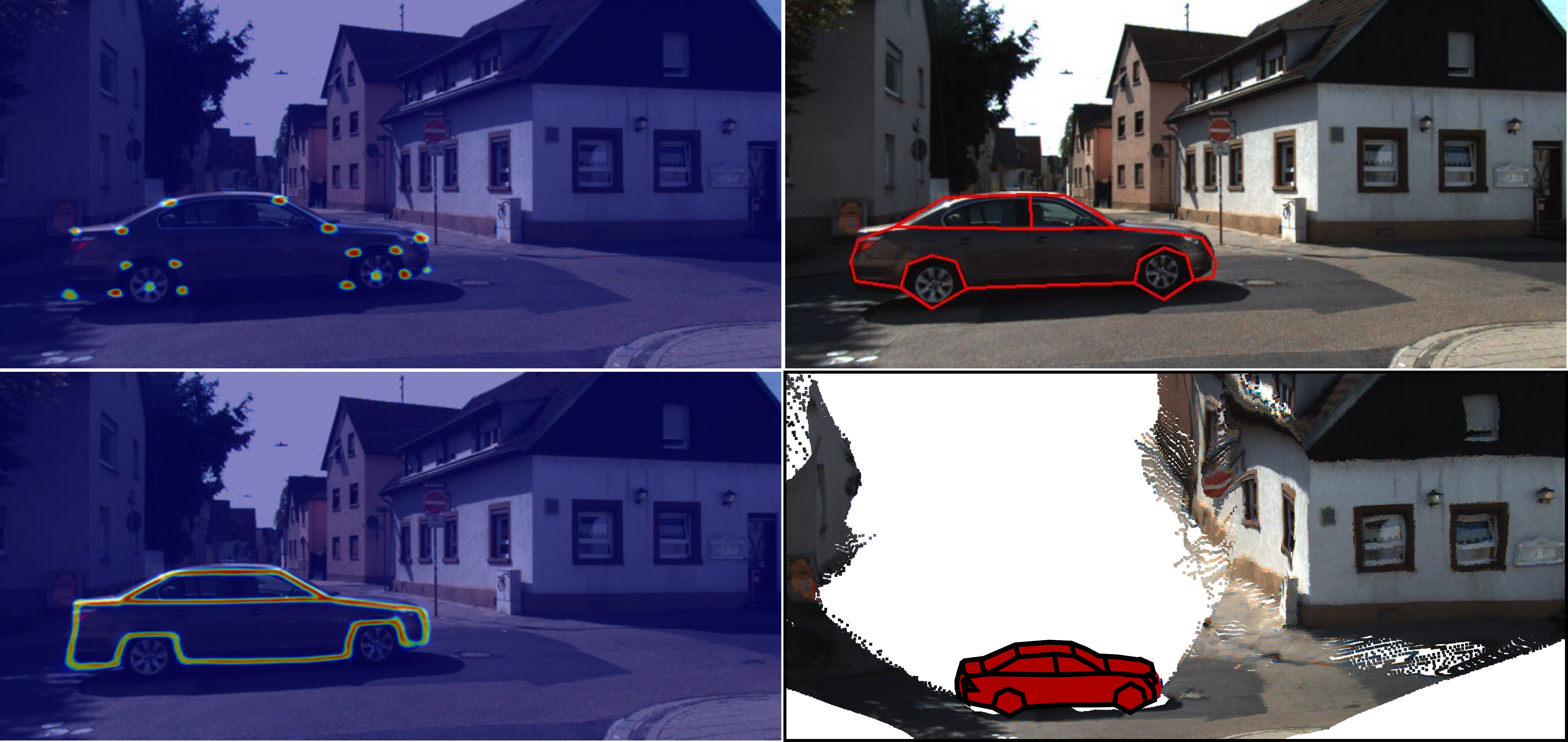}
\caption{Qualitative results of our method. Left: Heatmaps for vehicle keypoints (top) and vehicle wireframes (bottom) superimposed to the input image. Right: Input image and backprojected 3D wireframe of a reconstructed vehicle (top) and 3D view on the reconstructed scene (bottom).}
\label{fig:qualResults}
\end{figure}
The highly dynamic nature of street environments is one of the biggest challenges for autonomous driving applications. The precise reconstruction of moving objects, especially of other cars, are fundamental to ensure safe navigation and to enable applications such as interactive motion planning and collaborative positioning. To this end, cameras provide a cost-effective solution to deliver perceptive data of a vehicle's surroundings. 
However, the projection from 3D to 2D images leaves many ambiguities about 3D objects, causing the retrieval of their pose and shape to be ill-posed and difficult to solve. 
To confine the parameter space, deformable models can be used as shape prior and are aligned with the objects in the image to recover their pose and shape. 
In this paper, we make use of such a deformable vehicle model and present a method that fully reconstructs vehicles in 3D given street level stereo image pairs, allowing the derivation of precise 3D pose and shape parameters.
Earlier approaches used model edges and contours to align them with image edges, usually derived from gradient information \cite{Leotta2011, Payet, Ramnath2014}. However, these approaches are highly sensitive to illumination, reflections, contrast, object color and model initialisation, because these factors can cause erroneous edge-to-edge correspondences, thus prohibiting a correct model alignment. Instead, recent approaches leveraged keypoint detections to be used for model alignment \cite{KrishnaMurthy2018,Wenhao2018, KrishnaMurthy2017a,Pavlakos2017}. However, compared to edges, keypoints are less stable in the way that already small localisation errors are likely to cause large errors in 3D space. 
We start from the idea that if it is possible to detect keypoints, it should also be possible to detect model edges and contours without the explicit dependency on good image gradients. 
Based on initially detected vehicles, we make the following contributions in this paper:
\textbf{(1)} We propose a novel multi-task convolutional neural network (CNN) that simultaneously detects vehicle keypoints and vehicle wireframe edges (cf. Fig.~\ref{fig:qualResults}) and also outputs a probability distribution for the vehicle's orientation. For the orientation estimation, we expand the overlapping viewpoint class definition proposed in \cite{Mousavian2017} by defining a novel hierarchical class and classifier structure. We define a novel loss for the detection of keypoints and wireframes, which we believe allows the usage of a single encoder-decoder ("hourglass") CNN and makes the repeated hourglass architecture proposed in \cite{Newell2016, KrishnaMurthy2017} unnecessary. 
\textbf{(2)} Instead of relying on gradient based edge representations \cite{Coenen2019,Leotta2011,Ramnath2014}, which highly depend on illumination, contrast, and object color, we incorporate the wireframe predictions given by our CNN into our reconstruction approach. Since we train our CNN to distinguish between wireframe edges belonging to different sides of the vehicle, we not only avoid the problems due to low contrast of vehicle edges and silhouettes, but also achieve better prospects for edge-to-edge correspondences. 
\textbf{(3)} For the purpose of vehicle reconstruction, we build upon our previous probabilistic model and optimisation procedure \cite{Coenen2019} and significantly extend the model by adding state prior terms, leading to major improvements of the reconstruction results. 
To avoid the error source of incorrect keypoint localisations, in contrast to \cite{KrishnaMurthy2018,Pavlakos2017,KrishnaMurthy2017}, we build our probabilistic model directly on the raw keypoint heatmaps obtained by our CNN.
\textbf{(4)} We propose a new dataset for vehicle reconstruction, which exceeds the famous KITTI dataset \cite{KITTI} as it not only delivers 2D and 3D reference bounding boxes, but precisely fitted vehicle CAD models, which allow the evaluation of shape reconstructions and vehicle categorisation or identification.

%====================================================================================================
\section{Related Work}
One of the biggest challenges in image based reconstruction and pose estimation of vehicles is the enormous variability of appearance, caused by the intra-class variability of vehicles and by different viewpoints of the acquired images. One strategy  to overcome these problems is to train viewpoint and/or category specific classifiers \cite{Hodlmoser2013, Krause2013,Ozuysal} or to learn viewpoint specific shape templates \cite{Payet} to reason about vehicle pose and/or shape. However, in these approaches the viewpoint usually is discretised into a set of viewpoint bins and thus only coarse viewpoint estimates are delivered as output. 3D vehicle detection approaches such as \cite{3DOP} and \cite{Mousavian2017} deliver oriented bounding box estimates in 3D using CNNs. However, describing objects by a box only gives a very coarse representation of their shape.
Due to the ambiguous representation of 3D object information in images, the usage of 3D models as shape prior can be extremely beneficial for the task of object reconstruction. In this context, CAD vehicle models can be used directly to guide the vehicle reconstruction \cite{Displets}. However, finding the best suitable among the vast amount of existing CAD models becomes intractable quickly. Instead, deformable shape representations learned from a set of reference shapes, such as signed distance functions (SDF) \cite{Engelmann} or active shape models (ASM) \cite{Zia2} are more flexible representations to cope with the intra-class variability of vehicles. Using such prior models, a shape aware reconstruction is conducted, in which an instance of the deformed and transformed shape model is fitted to the observations to derive the target pose and shape parameters.
Matching a shape prior with image observations can be done via curve alignment of backprojected model edges and image edges \cite{Coenen2017, Leotta2011, Payet, Ramnath2014} or via alignment of the backprojected model silhouette with an instance segmentation mask in the image \cite{Dame, Kar2015,Prisacariu2012, Wang2019}. However, matching image edges highly depends on the illumination and contrast of the objects and on a good model initialisation to be able to establish correct edge-to-edge correspondences. On the other hand, silhouette matching inherits pose and shape ambiguities and neglects object details and structures. To counteract these problems, we define a vehicle wireframe that contains edges not only representing the silhouette but also boundaries between vehicle parts and learn a CNN to detect the wireframe, differentiating between edges belonging to different sides of the vehicle (left, right, front, back). This enables an illumination/contrast invariant, finer-grained and more robust establishment of correspondences for model fitting.

\cite{Coenen2017, Engelmann, StreetSideDet} propose methods for 3D vehicle reconstruction solely based on 3D points obtained from stereo images or laserscanning, respectively. In contrast to images, 3D points provide explicit 3D information, but when used as the only data source they deliver a rather sparse and incomplete representation of the object and valuable image cues are disregarded completely for model fitting.

Another strategy to shape-aware vehicle reconstruction is to match semantic model keypoints with their corresponding keypoints detected in the image. Traditionally, handcrafted features are used in \cite{Bao,Coenen2019,Li2011,Lin,Zia2} for the model to keypoint alignment, while recently CNNs are applied to detect keypoints \cite{KrishnaMurthy2018,DeepManta,Wenhao2018, KrishnaMurthy2017a,Pavlakos2017}. Typically, to infer the object pose and shape, the backprojection error of model keypoints and detected keypoints is minimized which makes these approaches prone to imprecise, incorrect and missing keypoint localisations. In contrast, we do not infer pointwise keypoint locations but instead build our model fitting approach on probabilistic keypoint heatmaps generated by our CNN.

End-to-end approaches that take the image as input and infer pose and shape of the target vehicle directly are presented in \cite{3DRCNN, Manhardt2019}. These methods require a vast amount of expensive training data, which is why often synthetically generated data is used for learning \cite{3DRCNN, Manhardt2019}. However, training CNNs on synthetic images usually leads to a drop of performance when applying the CNN to real world images.  Besides, explicit scene and/or model knowledge is disregarded in \cite{3DRCNN, Manhardt2019}. In contrast, we derive scene knowledge from the data and incorporate it as prior information.

%====================================================================================================
\section{Shape aware vehicle reconstruction}
\subsection{Overview}
\begin{figure}[ht]
\centering
\includegraphics[width=1.0\columnwidth]{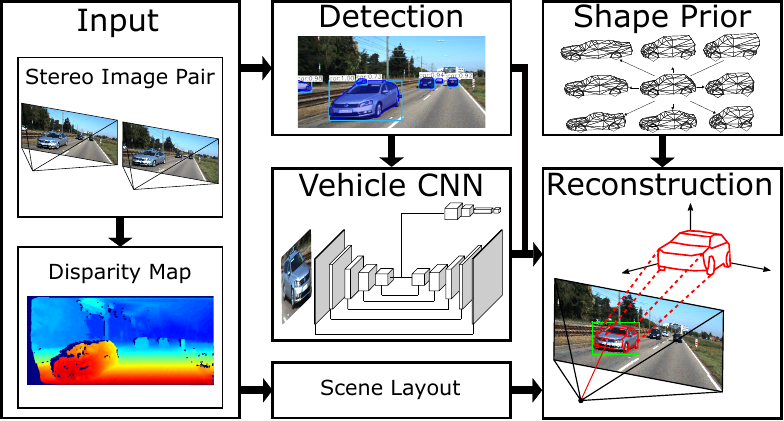}
\caption{Overview of our framework.}
\label{fig:overview}
\end{figure}

The goal of our method is to recover the precise pose (i.e. position and orientation) as well as the shape of vehicles detected from street-level stereo images. For this purpose, we fit a 3D vehicle model to the detected vehicles. We learn a deformable model as shape prior and formulate a probabilistic model to find the best fitting vehicle model by making use of a new vehicle CNN, trained to predict vehicle keypoints and wireframe edges, as well as the vehicle's viewpoint.
A schematic overview of our method is shown in Fig. \ref{fig:overview}.
The input to our method are stereo image pairs, incl. their interior and relative orientation parameters. To derive 3D information, we make use of the ELAS matcher \cite{ELAS} to calculate a dense disparity map for every stereo pair and reconstruct 3D points $\mathbf{X}$ via triangulation for every pixel of the reference image (the left stereo partner).
%***************************
%***************************
%______________________________________________
\subsubsection{Notation}
Given a set of detected vehicles $\mathbf{v}_k \in \mathcal{V}$, our goal is to associate each vehicle with its state vector $\mathbf{s}_k = (\mathbf{t}_k, \theta_k, \gamma_k)$, comprising its pose and shape parameters. After determining the ground plane $\Omega \in \mathbb{R}^3$, we describe the vehicle pose by its 2D position $\mathbf{t}_k$ on the ground plane and its heading $\theta_k$, i.e. the rotation angle about the normal vector of the ground plane; $\gamma_k$ is a vector of shape parameters determining the shape of the 3D deformable ASM representing each vehicle (cf. Sec. \ref{sec:ASM}).

%______________________________________________
\subsubsection{Scene layout} \label{sec:sceneLayout}
We use the stereo data to derive knowledge about the 3D layout of the scene, represented by the 3D ground plane and a probabilistic free-space grid map.

\textbf{Ground plane $\Omega$:} We apply RANSAC to the stereo point cloud $\mathbf{X}$ to find the ground plane $\Omega$ as plane of maximum support. All inliers of the final RANSAC consensus set are stored as ground points $\mathbf{X}_\Omega \subset \mathbf{X}$.
Requiring vehicles always to be located on the ground plane, we are able to fix three of the 6~DoF vehicle pose parameters (1 translational + 2 rotational parameters) and, thus, to constrain the parameter space of the model fitting approach.

\textbf{Probabilistic free-space grid map $\Phi$:}
Based on the ground plane points $\mathbf{X}_\Omega$ and all the points not belonging to the ground plane, thus representing arbitrary objects $\mathbf{X}_{Obj} = \mathbf{X} \setminus \mathbf{X}_\Omega$, it is possible to reason about free space, i.e. non occupied areas, in the observed scene. To represent the free space areas we create a probabilistic free space grid map $\Phi$. For this purpose, we create a grid in the ground plane consisting of square cells with a side length $l_\Phi$. For each grid cell $\Phi_g$ with $g\in[1,G]$ we count the number of ground points $n^g_{\Omega}$ and the number of object points $n^g_{Obj}$ whose vertical projection is within the respective cell. We define the probability $\rho_g$ of each cell to be free space as the ratio of $n^g_{\Omega}$ and the sum of $n^g_{\Omega}$ and $n^g_{Obj}$.
Grid cells without projected points are marked as \textit{unknown}.

%______________________________________________
\subsubsection{Vehicle detection}
To initially detect vehicles we apply the pretrained mask R-CNN (mRCNN) \cite{maskRCNN} to the reference image. Besides its good performance it has the advantage of not only delivering bounding boxes but also an instance segmentation mask for every vehicle. To obtain a list of $k \in [1,K]$ detected vehicles $\mathbf{v}_k = (\mathbf{X}_k, ^l\mathcal{B}_k, ^r\mathcal{B}_k)$, we associate each detection with the object points $\mathbf{X}_k$ reconstructed from the pixels belonging to the respective segmentation mask, as well as with its left and right image bounding boxes $^l\mathcal{B}_k$ and $^r\mathcal{B}_k$, the latter being derived from the dense stereo correspondences.

%______________________________________________
\subsubsection{Shape prior} \label{sec:ASM}
Similar to \cite{Zia2} we use a 3D ASM as vehicle shape prior. 
The ASM is learned by applying principal component analysis (PCA) to a set of manually annotated keypoints $\mathcal{K}$ of 3D CAD vehicle models. 
A deformed vehicle ASM is defined by the deformed vertex positions $\mathbf{\nu}(\gamma)$, which can be obtained by the linear combination
\begin{equation}
\mathbf{\nu}(\mathbf{\gamma}) = \mathbf{m} + \sum_{j} \gamma^{(j)} \ \sigma_j \  \mathbf{e}_j
\end{equation}
of the mean model $\mathbf{m}$ and the eigenvectors $\mathbf{e}_j$, weighted by the square root of their corresponding eigenvalues $\sigma_j^2$ and scaled by the object specific shape parameters $\gamma^{(j)}$. 
A fully parametrised instance of a 3D vehicle ASM on the ground plane, denoted by $M(\mathbf{s})$, can be created according to the state vector $\mathbf{s}$ by computing the deformed keypoints using the shape vector $\gamma$ and subsequently shifting and rotating the whole model on the ground plane according to the translation vector $\mathbf{t}$ and the heading angle $\theta$. 

\begin{figure}[ht]
\centering
		\includegraphics[width=1.0\columnwidth]{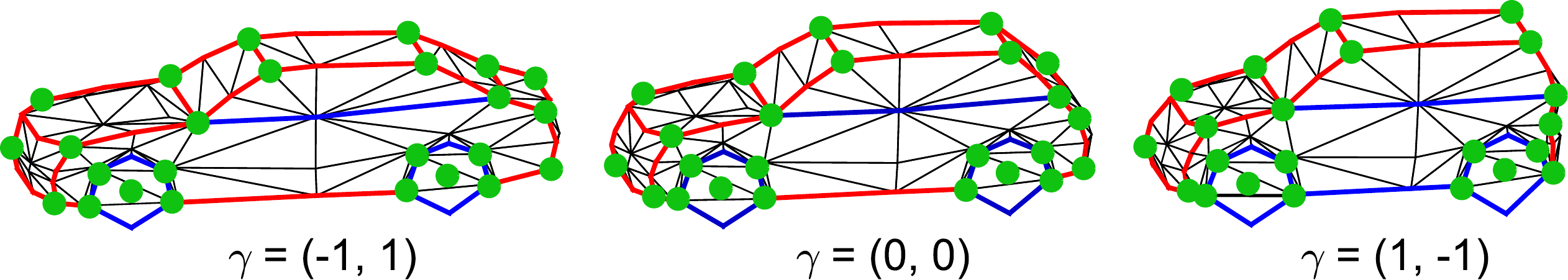}
	\caption{ASM: Triangulated surface (black), wireframe (red/blue: crease/semantic edges) and keypoints (green).}
\label{fig:ASM}
\end{figure}

\textbf{Geometrical representation:} We represent the model surface by defining a triangular mesh $M_{\Delta}$ for the ASM shape vertices $\mathcal{K}$. Further, we use a subset of $\mathcal{K}$ to define a wireframe $M_{\mathcal{W}}$ of the vehicle model, consisting of two types of edges: \textit{crease} edges that describe the outline of the vehicle and \textit{semantic} edges, describing the boundaries between semantically different vehicle parts. Another subset $\mathcal{K}_A \in \mathcal{K}$ are chosen as keypoints for which we learn an image based detector described later (cf. Sec.~\ref{sec:VehicleCNN}). Fig.~\ref{fig:ASM} shows the triangulations of several deformed models, their wireframes and the keypoints $\mathcal{K}_A$.
%========================================================================================
%

%====================================================================================================
\subsection{CNN structure} \label{sec:VehicleCNN}
\begin{figure*}[ht]
\centering
		\includegraphics[width=0.99\textwidth]{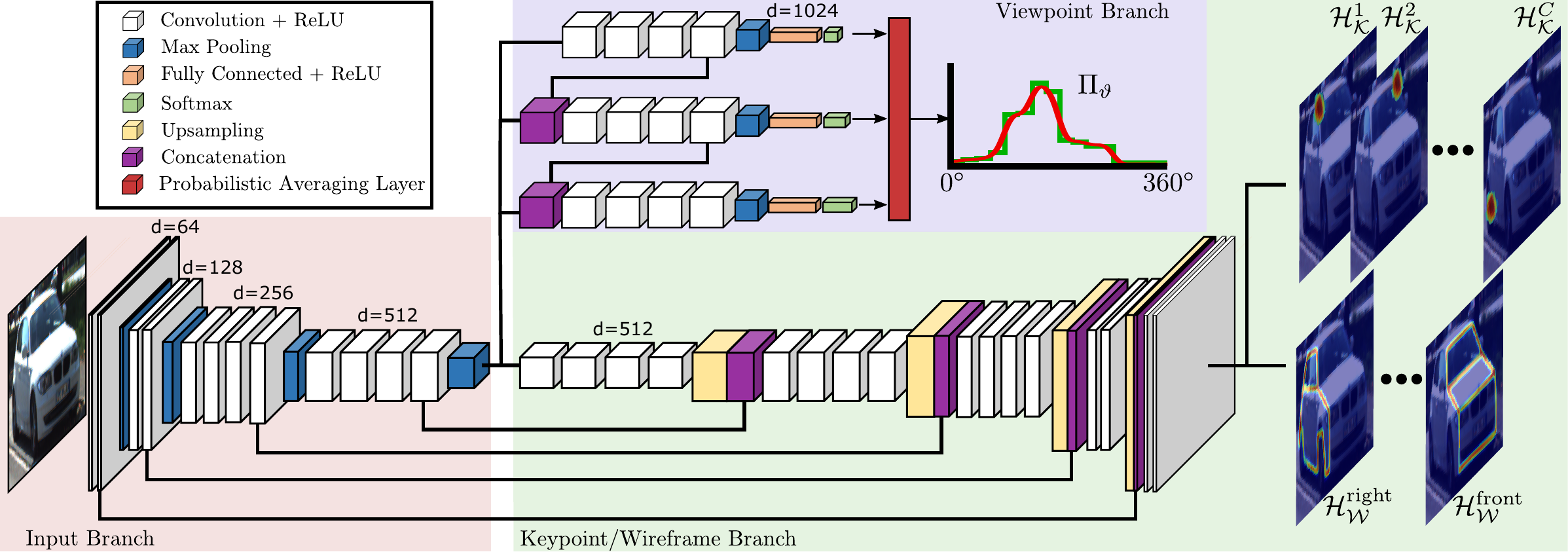}
	\caption{Architecture of our mulit-task CNN. The input is a 3 channel image of size 224x224. The convolutional filters have size 3x3, max pooling and upsampling use filter size 2x2 and stride 2. The number of filters is denoted by d in the figure.}
\label{fig:VehicleCNN}
\end{figure*}

Our multi-task CNN consists of one common input branch and two individual output branches, each of them corresponding to one task, respectively. The overall architecture of the network can be seen in Fig.~\ref{fig:VehicleCNN}.
The input to the network are images showing a vehicle, cropped by the bounding box. The network output consists of a probability distribution for the vehicle's viewpoint, a probability heatmap for each of a set of vehicle keypoints, and probability heatmaps for the vehicle wireframe edges.

\textbf{Input branch:} This branch contains a series of shared convolutions and max pooling layers, thus acting as a shared backbone feature extractor, adopting the architecture of the VGG19 network \cite{VGG19}.

\begin{figure}[H]
\centering
\subfloat[] {\includegraphics[width=0.5\columnwidth]{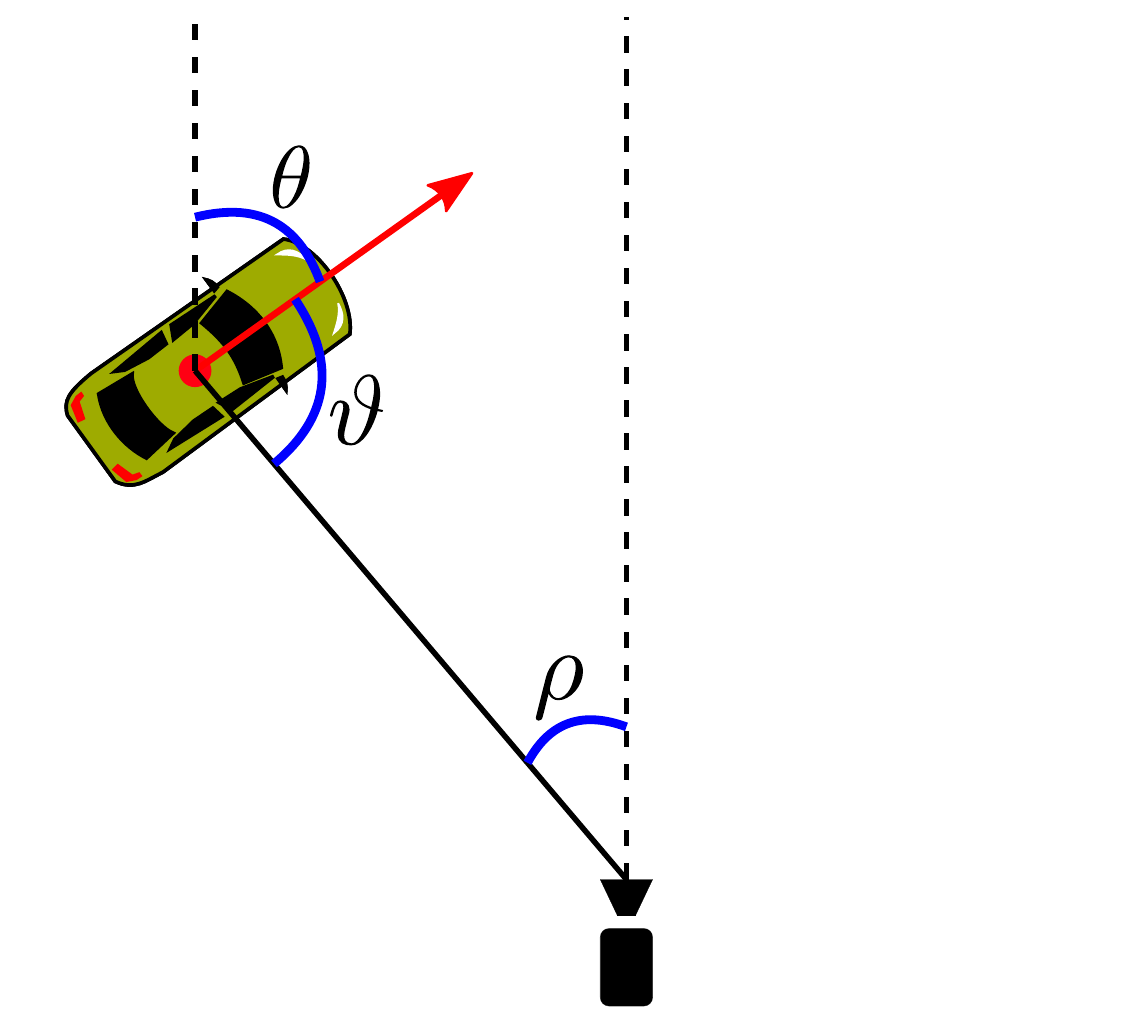}\label{fig:ViewpointDef}}
\subfloat[] {\includegraphics[width=0.5\columnwidth]{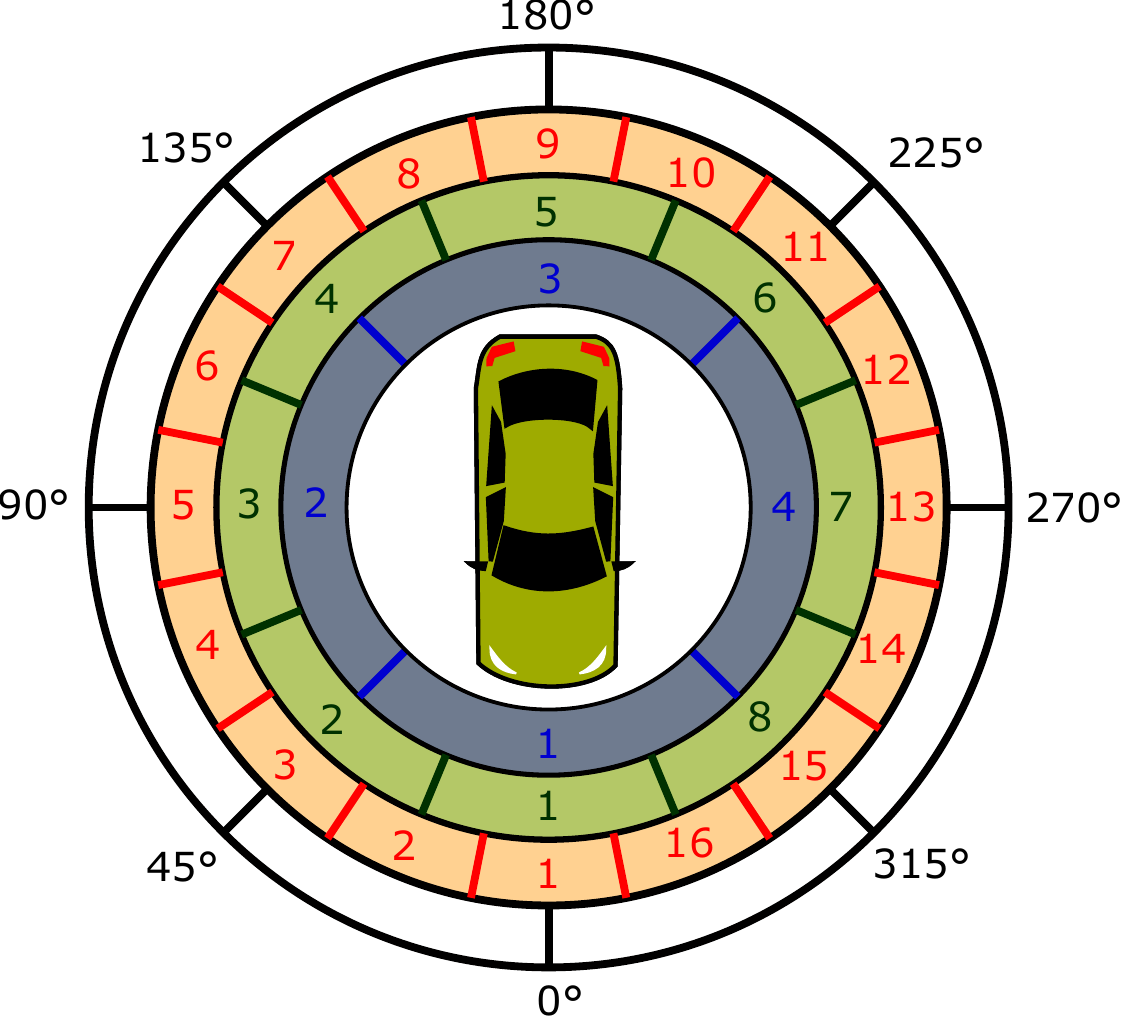}\label{fig:ViewpointClasses}}
\caption{Definition of the viewpoint angle (a) and of the hierarchical viewpoint classes in blue, green, and red (b).}
\label{fig:ViewpointBranch}
\end{figure}

\textbf{Viewpoint branch:} 
We design the viewpoint branch to output a probability distribution $\Pi_\vartheta$ for the vehicle viewpoint $\vartheta$, which describes the aspect under which the vehicle is seen. As depicted in Fig.~\ref{fig:ViewpointDef}, the viewpoint is defined as the angle between the image ray to the center of the vehicle and the vehicle's longitudinal axis (red arrow in Fig~\ref{fig:ViewpointDef}). Given the direction of the image ray $\rho$ the vehicle orientation can directly be computed from the viewpoint via $\vartheta = 180^\circ-\theta - \rho$.
In order to derive a probability distribution for the viewpoint of the vehicle, which can later be used in our probabilistic formulation for model fitting we set up a classification network rather than a regression network. The classes correspond to discretised orientation bins for the viewpoint estimation. 
The design of our viewpoint branch follows two assumptions regarding the general behavior of a viewpoint bin classifier: First, we expect the classification accuracy to decrease with an increasing number and therefore a finer definition of viewpoint bins. 
Second, we expect classification errors to primarily occur for vehicles with viewpoint angles close to the bin borders, i.e. being distributed close to the diagonal of the confusion matrix.
Inspired by \cite{Yan2015}, to consider the first assumption, we combine three hierarchical layers of classes for classification, containing different numbers (4, 8 and 16) of viewpoint classes. Thus, we believe to profit from the higher classification accuracy for the coarse class definition and the finer level of detail of the fine class definition. To consider the second assumption, we let the viewpoint bins of the individual layers overlap so that finer viewpoint classes do not share borders with coarser classes. Our hierarchical division of viewpoint classes can be seen in Fig.~\ref{fig:ViewpointClasses}. 
%========================================================================================
%
%
%========================================================================================
However, in contrast to \cite{Yan2015}, we do not apply conditional hierarchical classification, in which the output of the coarse classification layers decides about the execution of individual classifiers for the finer layers, because in that case, the output of images that are routed to an incorrect fine classifier cannot be corrected anymore.
Instead, we split the viewpoint branch into three independent softmax classification heads, one for each layer, and establish a skip-connection between features extracted at coarser layers to the feature extraction pipeline of the next finer layers (cf. Fig.~\ref{fig:VehicleCNN}). In this way, the fine category classifiers can profit from the information extracted by the coarse classifiers but are less dependent on the input.
The output of the classification heads is fed into a probabilistic averaging layer. Thus, our hierarchical class structure serves two purposes. On the one hand, we profit from the more reliable output of the coarse layers while still leveraging the more detailed output of the finer layers. On the other hand, due to the overlapping viewpoint bins, we mitigate the effect of misclassifications occuring between neighboring viewpoint classes.

%========================================================================================
%
%
%========================================================================================
%
\textbf{Keypoint/Wireframe branch:}
This branch corresponds to a decoder network, upsampling the output of the input branch to the original input resolution. Inspired by \cite{Newell2016} and \cite{KrishnaMurthy2017}, it is trained to produce one heatmap $\mathcal{H}^c_\mathcal{K}$ for every keypoint $c \in [1,C]$ in $\mathcal{K}_A$. Unlike previous work, we adapt the network to also output heatmaps $\mathcal{H}^w_\mathcal{W}$ for the vehicle wireframe edges. To this end, we subdivide the wireframe edges into four mutually non-exclusive subsets, each of which contains all edges belonging to the wireframe of one of the four vehicle sides $w \in \{\text{front}, \text{back}, \text{left}, \text{right}\}$. 
The values at each pixel position of the resulting heatmaps correspond to a probability for the presence of the respective keypoint/wireframe edge at that position. 
Together with the input branch, the keypoint/wireframe network follows a symmetrical UNet-like \cite{unet} architecture including skip connections between corresponding layers of the encoder and decoder blocks.
The head of this branch consists of $C+4$ binary classifiers using a sigmoid activation function to produce the $C$ keypoint and the four wireframe heatmaps. 

\textbf{Training:} 
The input branch is initialized from the corresponding layers of the VGG19 network \cite{VGG19}, pre-trained on ImageNet \cite{ImageNet} and frozen during training. The remaining convolutional layers are initialised using a normal distribution.
We train the viewpoint branch using categorical cross-entropy loss at each classification head given the groundtruth bin containing the groundtruth viewpoint of the training vehicle images.
We create training images for the keypoint/wireframe branch using annotated image keypoints. By automatically connecting keypoints corresponding to our wireframe definition we derive training data for the wireframe outputs. 
The training images are generated by placing 2D Gaussians at the corresponding keypoint positions, gaussian blurred edges along the reference wireframe, and zero everywhere else. The standard deviation of the Gaussians varies according to the distance of the vehicle to the camera and thus adapts to the size of the bounding box.
Training of the keypoint/wireframe branch is done by comparing the predicted heatmaps to the groundtruth heatmaps. In \cite{Newell2016} and \cite{KrishnaMurthy2017}, a mean squared error (MSE) loss is used for training, in which each pixel contributes equally to the loss. In our use case of detecting keypoints and edges, this leads to an unfavorably broad shape of the loss function due to the extremely small proportion of keypoint/wireframe pixels w.r.t. non-keypoint/non-wireframe pixels in the groundtruth. 
To overcome this problem, we apply a new custom keypoint/wireframe loss $\mathcal{L}_\mathcal{KW}$ with
\begin{equation}
\mathcal{L}_\mathcal{KW} = \text{MSE}_\text{true} + \text{MSE}_\text{false} + \text{MSE}_\text{pred}.
\end{equation}
Here, $\text{MSE}_\text{true}$, $\text{MSE}_\text{false}$ and $\text{MSE}_\text{pred}$ correspond to individual MSE, each computed for a different subset of pixels. In $\text{MSE}_\text{true}$ only pixels with groundtruth values larger than zero, whereas in $\text{MSE}_\text{false}$ only pixels with groundtruth values equal to zero are considered. In this way, the keypoint/wireframe and non-keypoint/non-wireframe pixels contribute to the loss with equal weight. Furthermore, we add an additional regularizing term $\text{MSE}_\text{pred}$ to the loss function In this term, only pixels  whose prediction exceeds a predefined threshold are considered. Thus, this term puts emphasis on keypoint/wireframe detections and acts as additional penalty of false positive outputs.
The network is trained using Keras \cite{Keras}, Adam optimizer for optimisation \cite{Adam2015}, a batch-size of 50, and a learning rate of $10^{-4}$. To improve training, we drop the learning rate by a factor of 10 after 5 validation accuracy plateaus, use Batch normalisation \cite{BatchNormal2015}, and apply Dropout \cite{Dropout} to the fully-connected layers with a rate of 0.5. 

%====================================================================================================
\subsection{Probabilistic Model}
Given the vehicle detections $\mathbf{v}_k$,
we fit a vehicle model $M(\mathbf{s}_k)$ to each detection by finding the optimal state variables $\hat{\mathbf{s}}_k = (\hat{\mathbf{t}}_k, \hat{\theta}_k, \hat{\gamma}_k)$. Neglecting the index $k$ (where it is possible) to simplify our notation in the following sections, $\hat{\mathbf{s}}$ can be derived by maximising the posterior
\begin{equation}
 p(\mathbf{s}|\mathbf{v}) = \frac{p(\mathbf{v}|\mathbf{s}) \cdot p(\mathbf{s})}{p(\mathbf{v})} \rightarrow \text{max}.
\end{equation}
Adapting the ideas from \cite{Coenen2019}, we further factorize the likelihood and the prior according to
\begin{equation}
p(\mathbf{s}|\mathbf{v}) \propto \underbrace{p(X|\mathbf{s}) \cdot p(\mathcal{H}_\mathcal{K}|\mathbf{s}) \cdot p(\mathcal{H}_\mathcal{W}|\mathbf{s})}_\text{Observation likelihood} \cdot \underbrace{p(\mathbf{t}) \cdot p(\theta)}_\text{State prior}. \label{eq:posterior}
\end{equation}
We minimise the negative logarithm of the posterior of Eq.~\ref{eq:posterior}. The individual likelihood and prior terms are explained in the following paragraphs.\\

%__________________________
\textbf{3D likelihood:} 
Based on the distances of the 3D points $X_k$ to the surface $M_\Delta$ of the model $M(\mathbf{s})$ we calculate the 3D likelihood as
\begin{equation}
\log p(X_k|\mathbf{s}) = -\frac{1}{P} \sum_{x\in X_k} \frac{d_{\sigma_x}(x, M_\Delta)}{2\sigma_x^2}.
\label{eq:E3D}
\end{equation}
Here, $P$ is the overall number of 3D points in $X_k$ and $\sigma_x$ is the depth uncertainty of the individual 3D point $x$. We apply the Huber norm to calculate $d_{\sigma_x}(x, M_\Delta)$, as it is more robust against outliers. This likelihood fits the 3D ASM to the 3D point cloud.\\

%__________________________
\textbf{Keypoint likelihood:}
To calculate this term we backproject the visible model keypoints of $\mathcal{K}_A$ to the stereo images, resulting in $c=[1,C_v]$ image points $\mathbf{u}_c^{l/r}$ for both, the left ($l$) and right ($r$) stereo images, respectively. The keypoint likelihood is calculated using the keypoint heatmaps $\mathcal{H}_\mathcal{K}$ by
\begin{equation}
\log p(\mathcal{H}_\mathcal{K}|\mathbf{s}) = -\frac{1}{2C_v} \sum_{i \in \{l,r\}} \sum_{c=1}^{C_v} \log \left(1- {^i}\mathcal{H}_\mathcal{K}^c(\mathbf{u}_c^i)\right)
\end{equation}
Here, $\mathcal{H}_\mathcal{K}^c(\mathbf{u}_c)$ denotes the output of the heatmap for the keypoint $c$ at the location $\mathbf{u}_c$. This term fits the 3D ASM to the predicted keypoints.\\

%__________________________
\textbf{Wireframe likelihood:}
This term is based on a measure of similarity between the backprojected edges of the model wireframe $M_\mathcal{W}(\mathbf{s})$ and the wireframe heatmaps $\mathcal{H}_\mathcal{W}$ resulting from our CNN. To this end, considering self-occlusion, we backproject the visible parts of the wireframe subsets $w \in \{\text{front}, \text{back}, \text{left}, \text{right}\}$ to the left and right images, resulting in binary wireframe images ${^l}I_\mathcal{W}^w$ and ${^r}I_\mathcal{W}^w$ with entries of 1 at pixels that are crossed by a wireframe edge and 0 everywhere else.
We blur the wireframe images using a Gaussion filter to account for generalisation effects. The size of the filter is defined according to the backprojection uncertainty of the model keypoints given the generalisation error of the ASM which is quantified to be 10~cm. The wireframe likelihood is calculated according to
\begin{equation}
\log p(\mathcal{H}_\mathcal{W}|\mathbf{s}) = -\frac{1}{2} \sum_{i \in \{l,r\}} \sum_w \log\left(1-BC({^i}I_\mathcal{W}^w, {^i}\mathcal{H}_\mathcal{W}^w)\right) 
\end{equation}
where we use the Bhattacharyya coefficient $BC(\cdot, \cdot)$ as similarity measure between the wireframe images and the wireframe heatmaps. This likelihood will be large if the backprojected wireframes correspond well to the wireframes predicted by the CNN.\\

%__________________________
\textbf{Position prior:}
The position prior is derived from the probabilistic free-space grid map $\Phi$ (cf. Sec~\ref{sec:sceneLayout}). It is calculated based on the amount of overlap between the minimum enclosing 2D bounding box $M_\mathcal{B}$ of the model $M(\mathbf{s})$ on the ground plane and the free-space grid map cells $\Phi_g$ given their probability $\rho_g$ of being free space:
\begin{equation}
\log p(\Phi|\mathbf{s}) = \frac{\lambda_\Phi}{A_\mathcal{B}} \sum_{g=1}^G \log(1-\rho_g) \cdot o(M_\mathcal{B}, \Phi_g).
\label{eq:Efree}
\end{equation}
$A_\mathcal{B}$ is the area of the model bounding box. The function $o(\cdot,\cdot)$ calculates the overlap between the model bounding box and a cell $\Phi_g$. 
The factor $\lambda_\Phi = \min(1, \frac{l_\phi}{\sigma_M})$ is used to weight this likelihood term based on the grid cell size $l_\Phi$ and the depth uncertainty $\sigma_M$ of a stereo-reconstructed point in the distance of the model $M(\mathbf{s})$. This likelihood penalises models that are partly or fully located in areas which are observed as not being occupied by 3D objects. \\

%__________________________
\textbf{Orientation prior:}
To calculate the orientation prior for the model $M(\mathbf{s})$ we use the probability distribution $\Pi_\vartheta$ for the vehicle viewpoint inferred by our CNN. We compute the viewpoint $\vartheta_M$ from the model orientation $\theta$. The image ray direction $\rho$ is derived from the ray connecting the camera projection center and the center of the vehicle model $M(\mathbf{s})$. The orientation prior is calculated according to
\begin{equation}
\log p(\theta) = \log \Pi_\vartheta(\vartheta_M) + \log \left(\frac{1+\cos(\vartheta_{\text{CNN}} - \vartheta_M)}{2}\right).
\end{equation}
$\Pi_\vartheta(\vartheta_M)$ denotes the probability for the angle $\vartheta_M$ according to the output of the viewpoint classification branch of the CNN. As we assume incorrect viewpoint classifications to appear especially between neighboring viewpoints, this term alone is prone to cause small orientation biases. This is why we additionally consider the cosine distance of the most likely viewpoint $\vartheta_{CNN}$ predicted from the vehicle CNN and the model viewpoint $\vartheta_M$ in this prior.\\

%====================================================================================================
\textbf{Inference}
To find the optimal pose and shape parameters for each detected vehicle we minimize the negative logarithm of Eq.~\ref{eq:posterior}. As this function is non-convex and discontinuous we apply the sequential Monte Carlo sampling approach described in \cite{Coenen2019} to approximate the parameter set for which the energy function becomes minimal. Starting from an initial state particle, we generate a number of particles in each iteration by jointly sampling the pose and shape parameters from a uniform distribution centered at the preceding parameter values. 
In contrast to \cite{Coenen2019}, the initial particle orientation is derived from the viewpoint estimated by the CNN. 
For more details we refer the reader to \cite{Coenen2019}.

%====================================================================================================
\section{Evaluation}
\subsection{Test data and test setup} \label{sec:setup}
We test and evaluate our proposed method on two datasets, the KITTI 3D object detection benchmark \cite{KITTI} and our own \textit{StereoVehicle} benchmark. 
The official KITTI evaluation metrics are designed to assess the joint performance of both, detection and pose estimation. As our approach only focuses on the latter and because we want to obtain further insights in the performance of our algorithm, we only use the KITTI training set with known annotations to evaluate our approach using own evaluation metrics. It consists of 7481 stereo iamges and provides the 3D object location and the orientation for every vehicle. It distinguishes three levels of difficulty (\textit{easy}, \textit{moderate} and \textit{hard}), which mainly depend on the level of object occlusion and truncation.

\subsubsection{StereoVehicle dataset}
For the acquisition of our dataset we equipped a vehicle with a calibrated and synchronised stereo camera rig using a baseline of 0.85\,m and recorded in total more than one hour of data during different day times in urban environments. Further details can be found in \cite{icsens2018}.
Compared to the KITTI benchmark \cite{KITTI}, our dataset has a larger image size (1936x1216), wider field of view using a focal length of 5\,mm, a significantly larger baseline, and a higher frame rate of 25\,fps.
To evaluate our approach we labeled 2289 vehicles in 1000 image pairs. In contrast to the KITTI dataset, which only delivers 2D and oriented 3D bounding boxes as references, we manually fitted the most similar model out of a large set of vehicle CAD models to the individual vehicles, and thus deliver the reference shape and the reference vehicle type in addition to its 3D pose. We distinguish between \textit{easy} vehicles, which are fully visible in the images, and \textit{moderate} vehicles, which are occluded or truncated. We intend to make the data publicly available.
 
%__________________________________________
\subsection{Parameter settings and training} 
We select the side length $l_\Phi$ of the free-space grid cells to be 25\,cm. 
For the number of the eigenvalues and eigenvectors to be considered in the ASM we choose $j \in [1,2]$, which we found to be a proper tradeoff between the complexity of the model and the quality of the model approximation.
For training our CNN we make use of the dataset provided by the authors of \cite{Zia3}, who labeled 36 different vehicle keypoints in a subset of images of the KITTI dataset. (Note that this subset is not used for evaluation.) We crop these images by the provided reference bounding boxes to use them to train our network. The viewpoint classes needed for the viewpoint branch are derived from the groundtruth viewpoint angles according to our viewpoint class definition in Fig.~\ref{fig:ViewpointBranch}.
To train the keypoint/wireframe branch, we make use of the 36 labeled landmarks and create the 2D reference heatmaps for the keypoints and wireframe edges as described in Sec.~\ref{sec:VehicleCNN}. We horizontally flip the training images to double the amount of training data by adapting the viewpoint classes and keypoint/wireframe labels accordingly. 
%__________________________________________
\subsection{Vehicle reconstruction results}
To evaluate the vehicle reconstruction, we compare the resulting pose parameters from each fitted 3D vehicle model and the reference data for location and orientation of the vehicles. We report results for position estimates whose euclidean distance from the reference position is smaller than 0.75\,m. To evaluate the orientation, we show results in three stages ($\theta_5$, $\theta_{10}$ and $\theta_{22.5}$), in which we consider an orientation to be correct if its difference from the reference is less than $5^{\circ}$, $10^{\circ}$ and $22.5^{\circ}$, respectively. Additionally, we calculate the average errors $\varepsilon_\mathbf{t}$ and $\varepsilon_\theta$ for position and orientation from all vehicle reconstructions. 

\subsubsection{Results on the KITTI benchmark}
The percentage of vehicles in the KITTI dataset detected by the mRCNN is shown in the leftmost column of Tab.~\ref{tab:Eval}. Besides, the resulting values for the described pose evaluation metrics are shown in the second rightmost column, compared to state-of-the-art results \cite{Coenen2019} reported in the rightmost column. Furthermore, we report results depending on the vehicle's distance from the camera. Additionally, we show the depth uncertainty $\sigma_x$ of a stereo reconstructed 3D point assuming an uncertainty of disparity of 1\,px.
Throughout the three levels of difficulty, we achieve a total percentage between 77.5 and 80.6\% of correct position estimates (cf. Tab.~\ref{tab:Eval}). However, it can be noticed that while the percentage of correct position estimates lies even between 96.3 and 97.1\% for vehicles having a distance between 5 and 10\,m, the amount decreases to 52.8\%-56.7\% for vehicles further away from the camera than 20\,m. Accordingly, the average error of position estimates also increases drastically and more than doubles from around 30\,cm for vehicles in a distance between 5 and 10\,m to more than 76\,cm for vehicles being more distant than 20\,m. We suspect the increasing depth uncertainty of distant 3D points to be responsible for this effect. 
Compared to the position estimates, the number of correct orientation estimates is higher with $\theta_{22.5}=  98.9\%$ and $\theta_5 = 86.7\%$ for the easy category (getting worse for the more challenging levels). However, the same effect of decreasing correct results for vehicles in increasing distance is visible for the orientation estimates, although less distinct compared to the position estimates. We also assume this effect to be caused by the increasing uncertainty with increasing distance to the camera.
Comparing our results to \cite{Coenen2019}, we obtain very similar values for the position estimates. However, we significantly outperform the orientation estimation results of \cite{Coenen2019}, especially for the moderate and hard levels by up to 5.7\% for the $\theta_5$ metric and even up to 9.4\% for the $\theta_{22.5}$ metric.
\begin{table}[ht]
\centering
\begin{footnotesize}
	\begin{tabular}{|c r|c c c c| c| c|}\hline
															& & \multicolumn{4}{c}{Vehicle distance} & &\\
															& & 5-10m & 10-15m& 15-20m & $>$20m & total& \cite{Coenen2019}\\  \hline
															& $\sigma_x [cm]$ &6-25 &25-58 &58-103 & $>$103 & &\\ \hline										
				\multirow{6}{0pt}{\rotatebox{90}{easy: 98.6\%}}
				&$\mathbf{t} [\%]$			& 96.8  & 94.8  & 76.0   & 52.8 & 79.4	 		& 80.8 	\\ 
				&$\varepsilon_\mathbf{t}$ [m] & 0.31 & 0.34 & 0.54 & 0.76 & 0.49		& - 		\\ 
				&$\theta_5 [\%]$ 			& 91.6  & 90.4  & 86.8   & 78.7 &  86.7 			& 84.8 	\\ 
				&$\theta_{10} [\%]$		& 99.4  & 98.4  & 96.9   & 94.4	& 97.1 				& 93.2	\\
				&$\theta_{22.5} [\%]$	& 99.7	& 99.6	& 98.8 	 & 97.6	&	98.9				& 94.8	\\
				&$\varepsilon_\theta$ [$^\circ$] & 2.75 & 2.87 & 4.32 & 5.94 & 4.02	& - 		\\	\hline	
				\multirow{6}{0pt}{\rotatebox{90}{moderate: 95.9\%}}
				&$\mathbf{t} [\%]$								& 		97.1	&	93.4	&	76.6	&	56.7	&	80.6	& 80.6	\\ 
				&$\varepsilon_\mathbf{t}$ [m] 		& 		0.30	&	0.37	&	0.55	&	0.76	&	0.50	& - 		\\ 
				&$\theta_5 [\%]$ 			& 								89.4	&	85.7	&	81.6	&	73.3	&	82.3	& 77.8	\\ 
				&$\theta_{10} [\%]$		& 								96.6	&	94.9	&	91.8	&	88.8	& 92.9	& 86.4	\\
				&$\theta_{22.5} [\%]$	& 								98.0	&	97.5	&	95.2	&	93.8	&	96.1	& 89.0	\\
				&$\varepsilon_\theta$ [$^\circ$] 	& 		4.49	&	4.88	&	7.35	&	9.43	&	6.6		& - 		\\ \hline	
				\multirow{6}{0pt}{\rotatebox{90}{hard: 85.5\%}}
				&$\mathbf{t} [\%]$								& 96.3	&	89.7	&	72.8		&	53.7	&	77.5	& 75.9		\\ 
				&$\varepsilon_\mathbf{t}$ [m] 		& 0.31	&	0.42	&	0.63		&	0.85	&	0.56	& - 			\\ 
				&$\theta_5 [\%]$ 			&   					85.7	&	78.6	&	74.9		&	67.4	&	76.1	& 70.4		\\ 
				&$\theta_{10} [\%]$		&   					93.0	&	88.3	&	85.1		&	81.9	& 86.7	& 78.4		\\
				&$\theta_{22.5} [\%]$	& 						95.9	&	92.8	&	89.3		&	87.1	& 91.0	&81.6			\\
				&$\varepsilon_\theta$ [$^\circ$] 	& 7.58	&	9.00	&	12.30		&	15.15	&	11.2	& - 			\\ \hline
	\end{tabular}			
	\end{footnotesize}
	\caption{Overall and distance dependent results of our vehicle reconstruction approach on the KITTI dataset.}
	\label{tab:Eval}
\end{table}
\vspace{-0.5cm}
\subsubsection{Results on the \textit{StereoVehicle} dataset}
The reconstruction results on our \textit{StereoVehicle} dataset are shown in Tab.~\ref{tab:Evalmapathon}. On the one hand, it is apparent that the overall results for the position estimates are distinctly better compared to the overall results on the KITTI dataset with 97.2\% for the easy and 92.8\% for the moderate level. Accordingly, the average position error achieved on our dataset is lower compared to the KITTI dataset. A reason for this might be the larger base length and consequently the lower depth uncertainty. However, the orientation estimates and average orientation errors are significantly worse compared to results achieved for the KITTI data, especially for the fine estimations reported in the $\theta_5$ metric. This effect can be caused by the domain gap of our CNN which is trained on KITTI data and therefore performs better on data from the same domain. Also, it has to be noted that the definition of our \textit{moderate} level covers the definitions of both, the moderate and hard level of the KITTI benchmark.
\begin{table}[ht]
\centering
\begin{footnotesize}
	\begin{tabular}{|c c c c c c c|}\hline
	& $\mathbf{t}$[\%] & $\varepsilon_\mathbf{t}$ [m] & $\theta_5$[\%] & $\theta_{10}$[\%] & $\theta_{22.5}$[\%] & $\varepsilon_\theta$ [$^\circ$] \\ \hline
	easy 			& 97.2 & 0.28 & 74.4 & 90.4 & 95.4 & 10.4 \\
	moderate 	& 92.8 & 0.35	& 70.1  & 85.8  & 90.8 & 16.8\\ \hline
	\end{tabular}			
	\end{footnotesize}
	\caption{Results of our vehicle reconstruction approach on our \textit{StereoVehicle} dataset.}
	\label{tab:Evalmapathon}
\end{table}

\begin{figure}[ht]
\centering
\includegraphics[width=0.9\columnwidth]{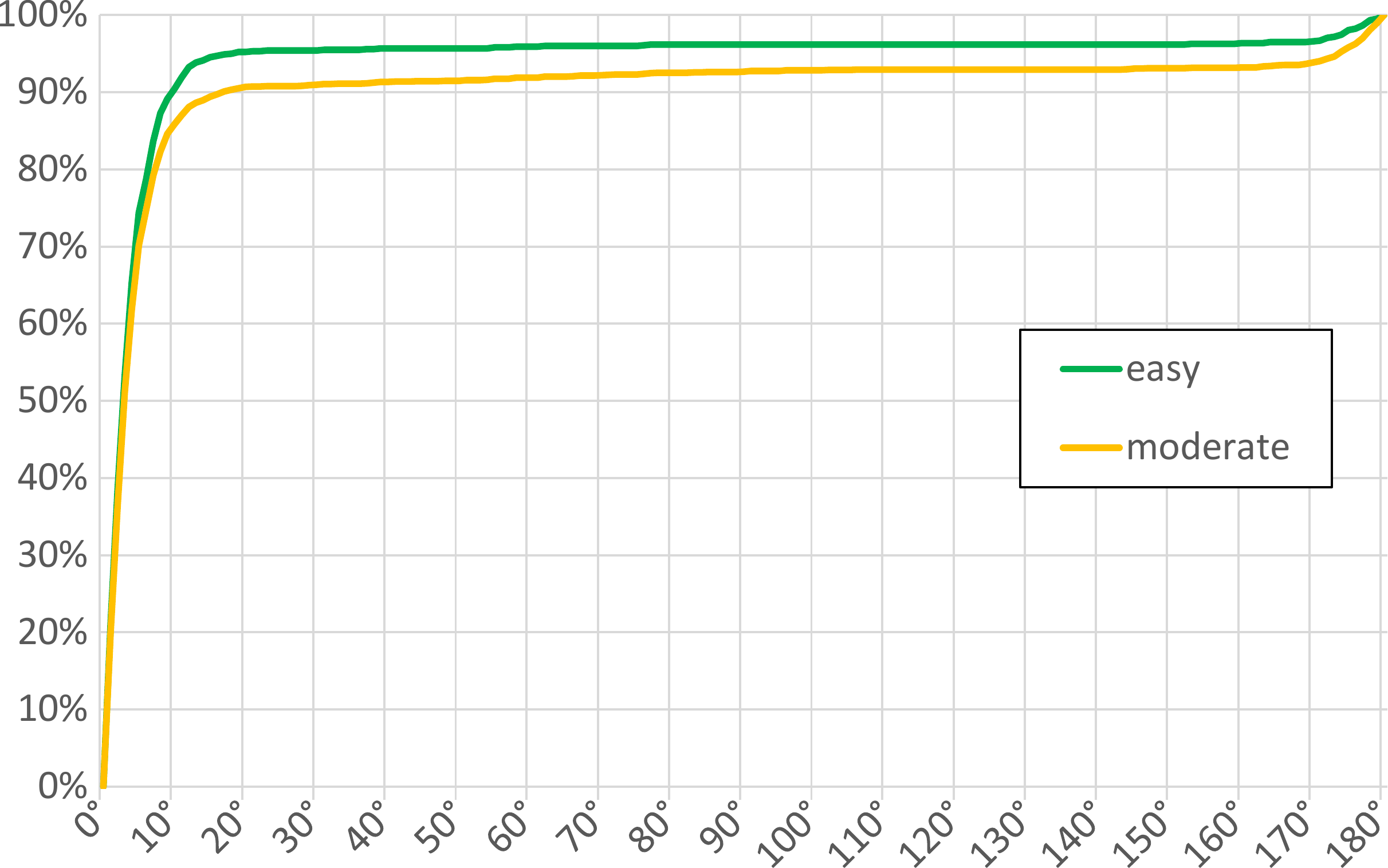}
\caption{Cumulative histogram of absolute orientation errors on our \textit{StereoVehicle} dataset.}
\label{fig:OriGraph}
\end{figure}

\vspace{-0.2cm} 
To obtain a closer look on the distribution of orientation estimation errors, we also show a cumulative histogram for the orientation errors resulting from our own dataset in Fig.~\ref{fig:OriGraph}. 
According to this, the majority of the incorrectly recovered orientations have an error between 170$^\circ$-180$^\circ$, i.e. they exhibit the opposed viewing direction and thus strongly influence the average orientation error. The reason for this effect might be ambiguities caused by the symmetric shape and appearance of vehicle front and back sides, which in some cases cannot be resolved by our approach.
%========================================================================================
%
%========================================================================================
\vspace{-0.3cm}
\section{Conclusion}
In this paper, we proposed a probabilistic approach for vehicle reconstruction, jointly incorporating 3D data, scene knowledge and predictions for vehicle orientation, keypoints and wireframes inferred by our proposed CNN. For the prediction of vehicle orientation we presented a novel hierarchical classification structure, allowing the derivation of a probability distribution to be used as prior. The CNN based detection of vehicle wireframe edges attenuates the dependency on good gradients. These innovations lead to state-of-the-art results on the KITTI object detection benchmark, as well as on our presented \textit{StereoVehicle} dataset, providing precisely fitted vehicle models as reference.

\section*{Acknowledgements}
This work was supported by the German Research Foundation (DFG) as a part of the Research Training Group i.c.sens [GRK2159].
%
%========================================================================================
%====================================================================================================
{\small
\bibliographystyle{ieee}
\bibliography{Literatur}
}

\end{document}